\documentclass[11pt]{article}

\usepackage[final]{acl}

\usepackage{times}
\usepackage{latexsym}

\usepackage[T1]{fontenc}
\usepackage[utf8]{inputenc}

\usepackage{microtype}

\usepackage{inconsolata}

\usepackage{graphicx}

\usepackage{booktabs}
\usepackage{xcolor}
\usepackage{amsmath}
\usepackage{amssymb}
\usepackage{algorithm}
\usepackage{algpseudocode}
\usepackage{tikz}
\usetikzlibrary{decorations.pathreplacing,arrows.meta,calc,positioning,fit,backgrounds}
\usepackage[most]{tcolorbox}

\definecolor{userc}{RGB}{30,90,170}
\definecolor{botc}{RGB}{30,130,70}
\definecolor{beepc}{RGB}{220,120,20}
\definecolor{kpc}{RGB}{0,140,140}
\definecolor{llmc}{RGB}{140,70,180}
\definecolor{termc}{RGB}{200,40,40}
\definecolor{startc}{RGB}{20,140,60}
\definecolor{stopc}{RGB}{190,60,60}
\definecolor{archproc}{RGB}{210,228,250}
\definecolor{archdet}{RGB}{253,236,200}
\definecolor{archcoord}{RGB}{205,238,210}
\definecolor{archterm}{RGB}{250,210,210}
\definecolor{archio}{RGB}{234,234,234}

\title{FEND: Far-End Detection in Outbound Voice AI Calls} 

\author{Abhishek Kumar \quad Suraj Agrawal \quad Aashraya Sachdeva \\
  Observe.AI, India \\
  \texttt{\{abhishek.kumar, suraj.agrawal, aashraya.sachdeva\}@observe.ai}}

\begin{document}
\maketitle

\begin{abstract}
When a conversational Voice AI agent places an outbound call, the far end may be a live person, a
voicemail greeting, an interactive voice response (IVR) system, or increasingly another AI system
answering on the callee's behalf. Standard Answering Machine Detection (AMD) reports only whether a human or a machine picked up, a design meant for human call centres, where one machine is as uninteresting as another. That single bit does not tell a Voice AI agent what to do,
since leaving a voicemail message, navigating an IVR and holding a short exchange with another AI
are three different actions and a two-valued label cannot distinguish them. In $1{,}000$ randomly sampled calls from our own outbound deployment, a live
person answered $48.1\%$, a voicemail system $44.6\%$, an AI $6.9\%$ and an IVR $0.4\%$. Our proposed framework, FEND, reads the far end's audio and its streaming transcript and runs one
detector for each automated class (voicemail, IVR, AI callee), with human the default when nothing
fires, returning a single label within $20$ seconds. On a public benchmark the voicemail components reach $100\%$ accuracy without ever misclassifying
a live human, where Asterisk's AMD misclassifies $30$ to $66\%$ of them. For the AI callee, acoustic features do not transfer across
recording conditions, but speech timing does, separating AI from human at $0.952$ ROC-AUC, because
synthesised speech leaves far shorter gaps between words. On production traffic, separating a live person from any automated far end reaches F1 $0.979$, and
only $2.9\%$ of live people are given an automated label, the error that matters most for an agent
acting on the result.
\end{abstract}

\section{Introduction}
\label{sec:intro}

An outbound call placed by a conversational Voice AI agent has to start by answering a question a
human caller answers without thinking: who, or what, just picked up? Deployed practice reduces this
to a single bit. Answering Machine Detection (AMD), the standard telephony service, reports whether
a human or a machine answered \citep{twilioamd,sinchamd,infobipamd}. That design comes from human
call centres, where the point was to avoid spending a live agent's time on a machine and one
machine was as uninteresting as another. When the caller is itself a voice AI agent the question changes.
Talk time is cheap, the expensive mistakes are hanging up on a person and reading a script to an
automated system, and the far end is no longer only a mailbox.

We therefore treat the far end as a four-class problem. \textsc{human} is a live person, and the
agent should proceed. \textsc{voicemail/am} is a mailbox greeting or an answering machine (AM), a
device or carrier service that records on the callee's behalf; the agent should leave its message
and release the line. \textsc{ivr} is automation that speaks for an \emph{organisation}, whether it
offers selectable options, plays a queue announcement or reads a recorded business greeting; the
agent should stay on the line. \textsc{ai callee} is an automated system that answers for an \emph{individual} and converses,
responding to what the calling agent says and possibly handing the call to a person; the agent
should introduce itself and hold a short exchange. Separating the last two by whom the automation
speaks for, rather than by whether options are offered, is what makes the four classes disjoint.
Merging any pair discards a decision the agent has to make, and merging \textsc{ai callee} into
\textsc{voicemail/am} is the costliest, because it converts a conversation that could still reach
the callee into a hang-up.

The classes are not equally served by existing work. We annotated $1{,}000$ calls sampled at random
from our own outbound deployment: a live person answered $48.1\%$, a voicemail system $44.6\%$, an
AI system answering on the callee's behalf $6.9\%$, and an IVR $0.4\%$. Consumer call screening is the reason the AI callee class exists at that scale, with handset assistants answering unknown numbers,
asking who is calling and why, and deciding whether to pass the call on \citep{twilio2025ios26}.

Our contributions are the following.
\vspace{-1mm}

\begin{itemize}\setlength{\itemsep}{2pt}\setlength{\topsep}{2pt}
  \item A four-class formulation of far-end detection with disjoint classes, replacing the human-or-machine label that telephony AMD returns, and a system, FEND, that implements it.
\vspace{-1mm}

  \item An evaluation on a public benchmark and on $1{,}000$ hand-annotated production calls, and
  against an open-source AMD implementation that can be replayed offline.
\vspace{-1mm}

  \item Evidence that the AI callee class is not separable from the far end's spectrum but is
  separable from its speech timing, together with the cross-corpus procedure that distinguishes a
  real signal from a channel artifact.
\end{itemize}
\vspace{-1mm}

\noindent To support
replication we provide the class definitions, the corpus splits, the detector prompts, and a
standalone reimplementation of Asterisk's AMD module that reproduces its decisions call for call
against Asterisk 20.6.0.
\vspace{-1mm}

\section{Related Work}
\label{sec:related}
\vspace{-1mm}

\paragraph{Answering machine detection.} Commercial platforms expose AMD as a configurable service
returning a human-or-machine label. Twilio detects in about four seconds with a tunable speed and accuracy trade-off
\citep{twilioamd,twilioamdfaq}, Sinch signals when the beep arrives \citep{sinchamd}, Infobip
reports the same two outcomes \citep{infobipamd}, and Telnyx adds a setting for Apple call
screening, though its output is still one class per call \citep{telnyxamd}. Cisco
\citep{ciscoamd} and CallFire \citep{callfireamd} describe pipelines over greeting length, the
silence after it, and tone rejection, a Vonage project detects the beep with a learned audio model
\citep{vonageamd}, and Asterisk's \texttt{app\_amd} is rule-based over silence and word-length
thresholds \citep{asteriskamd,minaj2020asterisk}. Research takes the same two-class view, from
energy and cadence heuristics \citep{gomez2014detector} to convolutional \citep{anisimov2017cnn}
and recurrent \citep{altwlkany2024rnn} models trained on contact-centre audio, and
\citet{saurav2026voicemail} detects voicemail in real time from fifteen temporal speech-activity
features at $96.1\%$ accuracy over $77{,}000$ production calls. A vendor evaluation of voicemail
detection in outbound agents reports the same limitation \citep{roark2026amd}. None of this work
distinguishes the kinds of machine, which is the distinction a voice AI agent needs.
\vspace{-2mm}

\paragraph{Detecting an AI on the line.} The natural instrument for the AI callee class is the
anti-spoofing literature, where ASVspoof~5 is the current large-scale benchmark
\citep{wang2025asvspoof5}. Two recent results bear on our setting: \citet{sharma2026voxenes} find
most of eight pretrained detectors near or below chance against current synthesis under
post-processing, and \citet{chandra2025deepfakeeval} report roughly a $48\%$ drop in audio detector area under the
curve on in-the-wild deepfakes. Our results agree with both. The other place to look is timing.
Consumer call screening is the immediate source of this class and vendors have begun documenting
how it interacts with AMD \citep{twilio2025ios26}, while work on turn-taking and endpointing
\citep{skantze2021turntaking,raux2008endpointing,lin2025fullduplexbench} treats conversational
timing as a control signal. We use timing instead as evidence about who is speaking, which to our
knowledge has not been done for far-end classification.
\vspace{-1mm}

\section{Methodology}
\label{sec:detectors}
\vspace{-1mm}

A far-end detector observes one channel of a live call, the side the callee is on, as a stream of
audio frames and of partial transcripts produced by automatic speech recognition (ASR). At every
ASR update it may emit a label, and the deployment commits to whichever label stands at a decision
deadline $W$ seconds after the call is answered. \textsc{human} is the default: a detector must
produce positive evidence to assign any other class, so a live person is mislabelled only when some
detector fires in error.

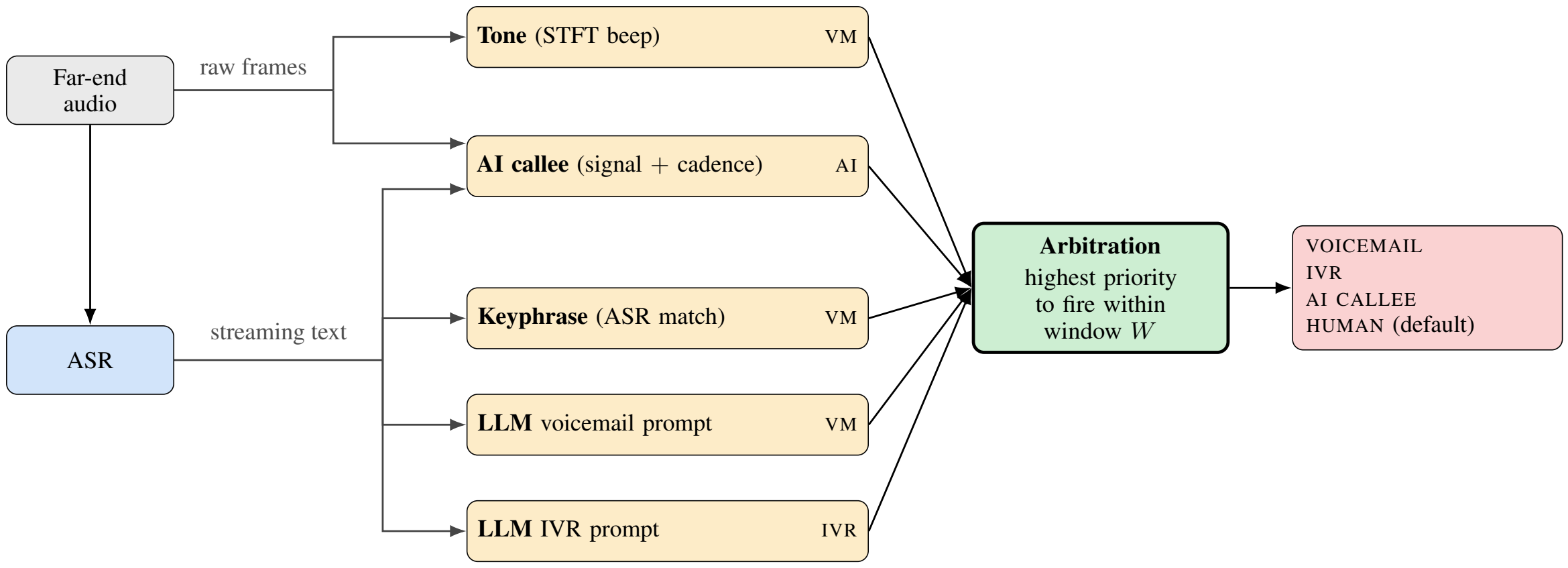
\begin{figure*}[t]
  \centering
  \resizebox{0.96\textwidth}{!}{%
  \begin{tikzpicture}[font=\small,>=Latex,
    src/.style={draw,rounded corners,fill=archio,align=center,minimum height=9mm,minimum width=22mm,inner sep=3pt},
    proc/.style={draw,rounded corners,fill=archproc,align=center,minimum height=9mm,minimum width=22mm,inner sep=3pt},
    comp/.style={draw,rounded corners,fill=archdet,align=left,minimum height=8mm,inner sep=4pt,text width=5.0cm},
    arb/.style={draw,very thick,rounded corners,fill=archcoord,align=center,inner sep=5pt,text width=3.0cm},
    res/.style={draw,rounded corners,fill=archterm,align=left,inner sep=5pt,text width=3.2cm},
    flow/.style={->,line width=0.7pt},
    tap/.style={->,line width=0.7pt,gray!55!black},
    bus/.style={line width=0.7pt,gray!55!black},
    lbl/.style={gray!60!black,font=\footnotesize},
  ]
  \node[comp] (c1) at (7.6, 3.00) {\textbf{Tone} {\footnotesize(STFT beep)}\hfill\textsc{vm}};
  \node[comp] (c2) at (7.6, 1.30) {\textbf{AI callee} {\footnotesize(signal $+$ cadence)}\hfill\textsc{ai}};
  \node[comp] (c3) at (7.6,-0.70) {\textbf{Keyphrase} {\footnotesize(ASR match)}\hfill\textsc{vm}};
  \node[comp] (c4) at (7.6,-2.10) {\textbf{LLM} {\footnotesize voicemail prompt}\hfill\textsc{vm}};
  \node[comp] (c5) at (7.6,-3.50) {\textbf{LLM} {\footnotesize IVR prompt}\hfill\textsc{ivr}};

  \path (c1.west)               -- ([yshift=3mm]c2.west) coordinate[midway] (amid);
  \path ([yshift=-3mm]c2.west) -- (c5.west)              coordinate[midway] (tmid);

  \node[src]  (audio) at (0,0 |- amid) {Far-end\\audio};
  \node[proc] (asr)   at (0,0 |- tmid) {ASR};
  \draw[flow] (audio.south) -- (asr.north);

  \draw[bus] (audio.east)      -- (3.20,0 |- audio);
  \draw[tap] (3.20,0 |- audio) |- (c1.west);
  \draw[tap] (3.20,0 |- audio) |- ([yshift=3mm]c2.west);
  \draw[bus] (asr.east)        -- (3.85,0 |- asr);
  \draw[tap] (3.85,0 |- asr)   |- ([yshift=-3mm]c2.west);
  \draw[tap] (3.85,0 |- asr)   |- (c3.west);
  \draw[tap] (3.85,0 |- asr)   |- (c4.west);
  \draw[tap] (3.85,0 |- asr)   |- (c5.west);

  \node[lbl,anchor=south,yshift=2pt] at (2.15,0 |- audio) {raw frames};
  \node[lbl,anchor=south,yshift=2pt] at (2.48,0 |- asr)   {streaming text};

  \node[arb] (arb) at (13.3,-0.3)
    {\textbf{Arbitration}\\[2pt]
     \footnotesize highest priority\\ to fire within\\ window $W$};
  \foreach \n in {c1,c2,c3,c4,c5}{\draw[flow] (\n.east) -- (arb.west);}

  \node[res] (resbox) at (17.6,-0.3)
    {\textsc{voicemail}\\ \textsc{ivr}\\ \textsc{ai callee}\\ \textsc{human} {\footnotesize(default)}};
  \draw[flow] (arb) -- (resbox);
  \end{tikzpicture}%
  }
  \vspace{-3mm}
  \caption{Detection architecture. Four detectors read the raw audio frames and the streaming
transcript and run concurrently, each able to fire one class. Firings accumulate over a window of
$W$ seconds and the highest-priority class that has fired wins, with \textsc{human} the default
when nothing fires. Section~\ref{sec:cascade} sweeps $W$ and the ordering.}
  \label{fig:arch}
  \vspace{-4mm}
\end{figure*}

No single model sees all four classes, because the evidence that identifies them is of different
kinds and arrives at different times. FEND therefore runs one independent detector per kind of evidence over two taps on the same call,
the raw audio frames and the streaming transcript, each emitting a binary firing for a single
class (Figure~\ref{fig:arch}). Each keeps running after it has fired, because the far end can
change during a call when a screener hands over to a person or a greeting ends in a beep. Two of them are tool calls on the large language
model (LLM) that is already running the dialogue, so they cost no extra model call.

\paragraph{Tone detector.} Looks for the record-now beep that follows a mailbox greeting, using a
short-time Fourier transform over the incoming audio frames (Appendix~\ref{app:beep}). It is the
only detector that does not depend on the ASR, so it is also the only one that still works
when the greeting is in a language the ASR handles poorly.
\vspace{-2mm}

\paragraph{Keyphrase detector.} A multilingual list of voicemail phrases, such as
\emph{``after the tone''} or \emph{``please leave your name''}, matched against the normalised
running transcript. It is precise, cheap, and trivially extended to a new language, but it only
fires on wordings someone has already added to the list (Appendix~\ref{app:keyphrases}).

\vspace{-2mm}

\paragraph{Voicemail LLM detector.} An LLM tool call that fires when the transcript shows an automated
mailbox system answering for the callee (Appendix~\ref{app:amdprompt}).

\vspace{-2mm}

\paragraph{IVR LLM detector.} A second tool with its own prompt, which fires on automation speaking for an organisation, such as
menus, queue announcements and recorded business greetings (Appendix~\ref{app:ivrprompt}).

\vspace{-2mm}

\paragraph{AI callee detector.} Measures \emph{how} the far end speaks rather than what it says, since an assistant answering for
an individual replies to whatever the voice AI agent says and cannot be recognised from a fixed
script. The ASR already reports when each word starts and ends, and from the far end's first turn we
summarise the pauses between words and how fast it speaks. A person pauses to think and to breathe, while a synthesised turn runs on. The detector is a
logistic-regression classifier over seven such timing summaries, with its threshold set on the
training half at a $5\%$ false-positive budget against live humans (Appendix~\ref{app:cadence}).

\vspace{-2mm}

\paragraph{Combining the detectors.} Detectors run concurrently and more than one can fire on the
same call, so a combination rule is needed. Two choices define it: the deadline $W$, and what to do
when two detectors fire inside it. We accumulate firings over the window and return the
highest-priority class that has fired by the end, ordered voicemail, IVR, AI callee, with human as
the default. Priority is by how conclusive the evidence is, so the class whose evidence is hardest
to produce by accident wins, which is not the same as the class that fires first. We treat both $W$
and the ordering as parameters and sweep them in Section~\ref{sec:inhouse} instead of fixing them
by assumption.

\vspace{-2mm}

\section{Experiments}
\label{sec:setup}
\vspace{-2mm}

\subsection{Datasets}
\label{sec:data}\label{sec:datasets}\label{sec:evalmethod}

\paragraph{Telephony-AMD (public).} $400$ labelled far-end clips across English, Spanish, French and
German, with four classes: \textsc{human}, \textsc{voicemail}, \textsc{answering machine} and
\textsc{ivr} \citep{telephonyamd2024}. We split it once, stratified by class and language, into a $119$-clip tuning half and a $281$-clip
evaluation half. Prompts and thresholds are fixed on the
tuning half before the evaluation half is scored.

\vspace{-2mm}

\paragraph{In-house Voice AI calls.} $1{,}000$ outbound calls sampled uniformly at random from our
deployment, annotated manually with far-end ground truth. Because the draw is
random, the counts estimate the mix the deployment encounters: \textsc{human} $481$ ($48.1\%$),
voicemail or answering machine $446$ ($44.6\%$), AI callee $69$ ($6.9\%$) and \textsc{ivr} $4$
($0.4\%$). Half the sample fits the one detector trained in-house and all in-house results are reported on
the other half. These calls contain customer content and are not released. Two forces are
enlarging the AI callee share. Handset vendors are rolling out call screening to consumers,
placing an assistant in front of a large share of personal numbers, and disclosure rules such as
EU AI Act Article~50 \citep{euaiact2024} and California's B.O.T. Act \citep{botact2018} require
automated callers to identify themselves, which turns part of the class into one that announces
itself.

\vspace{-1mm}

\paragraph{Corpora for the acoustic analysis.} The acoustic test of
Section~\ref{sec:aicallee} needs speech whose origin we know, so we assemble four sets and pass them
all through the same channel simulation and the same fixed-length window. Two are human speech from
independent sources, recorded bank calls \citep{wu2021harpervalley} and role-played call-centre
dialogues across fourteen English accents \citep{apptek2024}. One is public synthesised speech
\citep{chen2024voicebench}, the positive class. The fourth is a public robocall corpus
\citep{wspr2023robocallaudio}, which acts as a control: robocalls are mostly pre-recorded
\emph{human} speech, so they show whether a detector is responding to synthesis or merely to audio
that is not a live two-way call (Appendix~\ref{app:datasets}).

\vspace{-1mm}

\subsection{Models}
\vspace{-1mm}

Two different recognisers are involved, one per corpus. On the public Telephony-AMD clips we run Whisper \texttt{small}
\citep{radford2023whisper} in streaming mode with SimulStreaming
\citep{machacek2025simulstreaming,machacek2023whisperstreaming}, so that the public numbers are
reproducible. The in-house calls are transcribed by the commercial ASR the
deployment already runs, which we cannot disclose. The cadence features of Section~\ref{sec:aicallee}
are read from its word timings, so the timing result is tied to that recogniser. Both LLM detectors use GPT-4o \citep{openai2024gpt4o} 
and the prompts are frozen before evaluation (Appendices~\ref{app:amdprompt}
and~\ref{app:ivrprompt}).

\vspace{-1mm}

\subsection{Metrics and Evaluation}
\vspace{-1mm}

We report standard per-class precision, recall and F1, and for the ranking detectors the area
under the receiver operating characteristic curve (ROC-AUC) together with recall at a fixed budget
on human false positives. The most costly error is giving a live person an automated label, because the agent then ends a
conversation with a real human. We therefore treat the human false-positive rate (HFPR), the
fraction of true \textsc{human} far ends given any other label, as a constraint and not as one
metric among several. For the ranking detectors we fix an HFPR budget of $1$--$5\%$ and report the
recall reached inside it, because a detector with a high ROC-AUC but almost no recall inside that
budget cannot be deployed.

One evaluation choice is specific to the acoustic study. A measurement that separates two corpora
of \emph{human} speech from each other cannot be evidence about synthesis, so we calibrate each
candidate on human data alone before any human-versus-machine comparison. For values $S$ on one human corpus and $R$ on a second, independent one, we admit the measurement
only if the median difference, scaled by the interquartile range (IQR) of the reference corpus,
stays small:

\vspace{-6mm}
\begin{equation*}
|g| \le 1, \qquad g=\frac{\operatorname{med}(S)-\operatorname{med}(R)}{\operatorname{IQR}(R)} .
\end{equation*}

\vspace{-2mm}

Four of our five candidate measurements fail this null calibration (Appendix~\ref{app:signal}).

The remaining protocol follows the datasets. On Telephony-AMD, thresholds and prompts are frozen on
the $119$-clip tuning half and the $281$-clip evaluation half is scored once, with positive $=$
\textsc{voicemail}$+$\textsc{am} and \textsc{ivr} held out for the voicemail experiment and
positive $=$ \textsc{ivr} for the IVR experiment. The baseline is our offline port of Asterisk \texttt{app\_amd} at shipped defaults and at the best
of $1{,}152$ grid-searched configurations (Appendix~\ref{app:asterisk}). The timing and four-way
experiments use the in-house calls, split by source call so that no call appears in both halves,
fitted on one half and reported on the other at the deployment's own class distribution, with
$200$ resplits giving the reported intervals.

\vspace{-1mm}

\section{Results}
\label{sec:results}\label{sec:experiments}
\vspace{-1mm}

\subsection{Voicemail and Answering Machine}
\label{sec:voicemail}\label{sec:results-amd}\label{sec:baselines}
\vspace{-1mm}

Table~\ref{tab:results} reports the three voicemail detectors and their combination on both splits
of Telephony-AMD, with Asterisk for comparison. The key property holds on both splits: no detector
fires on any human clip, so precision is $1.000$ throughout. On the evaluation half the combination
reaches $100\%$ accuracy at $1.000$ recall. The tone detector is stronger on answering
machines ($96.9\%$ against $81.9\%$ on voicemail greetings), the LLM is the reverse ($100.0\%$
against $98.5\%$), and the keyphrase list is by far the most conservative. 

\begin{table}[t]
  \centering
  \small
  \setlength{\tabcolsep}{3pt}
  \begin{tabular}{@{}lcccccc@{}}
    \toprule
    & \multicolumn{4}{c}{\textbf{Fire rate (\%)}} & \multicolumn{2}{c}{\textbf{Binary}} \\
    \cmidrule(lr){2-5}\cmidrule(lr){6-7}
    \textbf{Component} & \textbf{VM} & \textbf{AM} & \textbf{Hum} & \textbf{IVR} & \textbf{Acc} & \textbf{Rec} \\
    \midrule
    \multicolumn{7}{@{}l}{\textit{tuning half} ($n{=}90$ scored, \textsc{ivr} held out)}\\
    Tone (beep)        & 70.0 & 100.0 & 0.0 & 0.0 & 0.900 & 0.842 \\
    Keyphrase          & 50.0 & 25.9  & 0.0 & 0.0 & 0.611 & 0.386 \\
    LLM (voicemail)    & 96.7 & 100.0 & 0.0 & 0.0 & 0.989 & 0.982 \\
    \textbf{Fusion}    & 96.7 & 100.0 & 0.0 & 0.0 & \textbf{0.989} & 0.982 \\
    \midrule
    \multicolumn{7}{@{}l}{\textit{evaluation half} ($n{=}215$ scored, \textsc{ivr} held out)}\\
    Tone (beep)        & 81.9 & 96.9 & 0.0 & 0.0 & 0.930 & 0.891 \\
    Keyphrase          & 47.2 & 36.9 & 0.0 & 0.0 & 0.633 & 0.423 \\
    LLM (voicemail)    & 100.0 & 98.5 & 0.0 & 0.0 & 0.995 & 0.993 \\
    \textbf{Fusion}    & 100.0 & 100.0 & 0.0 & 0.0 & \textbf{1.000} & 1.000 \\
    \midrule
    \multicolumn{7}{@{}l}{\textit{Asterisk} \texttt{app\_amd}\textit{, all 400 clips}}\\
    ~~shipped defaults & 93.1 & 98.9 & 65.8 & 92.4 & 0.731 & 0.959 \\
    ~~tuned (5-fold CV)& 100.0& 98.9 & 29.8 & 100.0& 0.886 & 0.995 \\
    \bottomrule
  \end{tabular}
  \vspace{-2mm}
  \caption{Voicemail components on Telephony-AMD. Binary metrics take positive $=$ VM$+$AM with
  \textsc{ivr} held out. Precision is $1.000$ for every component row: none fires on any human
  clip. The \texttt{app\_amd} human fire rate is the column that matters, and it is
  why a detector built for human call centres cannot be reused here.}
  \label{tab:results}
  \vspace{-4mm}
\end{table}

\paragraph{The open-source baseline.} We ported Asterisk \texttt{app\_amd} to an offline harness and
validated it clip-for-clip against Asterisk 20.6.0. With shipped defaults it labels $65.8\%$ of live humans as machines. Tuned by grid
search over $1{,}152$ configurations it still labels $29.8\%$, and it fires on every IVR clip (Appendix~\ref{app:asterisk}). The thresholds are set for a far end that answers with a short \emph{``Hello?''}, so a live person
who says more than a word or two runs past the greeting-length and word-count limits and is
declared a machine.

\vspace{-1mm}

\subsection{IVR}
\label{sec:ivr}\label{sec:results-ivr}
\vspace{-1mm}

On the held-out split the IVR detector reaches $93.8\%$ recall at $100\%$ precision in two of
three runs, and never fires on a human clip on either split. Two choices of definition produced most of
that number. First, an IVR is not only a menu: queue announcements such as \emph{``your call will be answered
in the order it was received''} and recorded business-hours greetings are automated far ends that
offer no option to select, and admitting them raised recall from $60.7\%$ to $96.4\%$ on the
tuning half. Second, breadth of that kind has a cost once a fourth class exists, because an
automated far end that is not a mailbox also describes an assistant answering for a person. We
therefore scope the class by whom the automation speaks for: an IVR answers for an organisation.
With that scoping the detector fires on none of the AI callee calls in our production sample, and
the public benchmark is unchanged (Appendix~\ref{app:ivrprompt}).

\subsection{AI Callee}
\label{sec:aicallee}

This is the class where content does not help. Acoustic evidence does not carry it: of five
measurements of the far-end signal, four fail the null calibration of Section~\ref{sec:setup}
because they track how much of the window contains speech rather than who produced it, and the
survivor reaches only $0.624$ ROC-AUC with zero recall at a $1\%$ budget. Repeating the comparison
inside a single recording channel appears to reach $0.858$, but one high-frequency energy
measurement reaches $0.866$ alone and gives $0.512$ on a voicemail control. We do not report it as a detection result
(Appendix~\ref{app:signal}).

Speech timing does carry it. Word timings survive any codec, and using only how the far end's first
turn is paced, a model reaches ROC-AUC $0.952$ $[0.902, 0.990]$ on held-out calls with recall
$0.774$ at a $5\%$ budget (Table~\ref{tab:timing}). A person pauses to think and to breathe, opening gaps between words, while a synthesised turn runs
on and is delivered faster. Table~\ref{tab:timing} lists the four most informative of the seven features the model uses, all
read from our deployment's ASR word timings. Two controls bound the claim: the same model separates AI from voicemail at $0.895$, so it is not
merely detecting pre-recorded audio, and the voicemail control sits at $0.707$. Only $62$ of the $481$ human far ends open with at least four timed words, so the
comparison is against the more talkative humans, and the positives appear to come from a single
screening product. In $17$ of the $69$ AI callee calls in the sample the far end answers our agent by name,
replying to its self-introduction with a phrase built around it, which requires processing speech
online and is something no mailbox greeting can do.

\begin{table}[t]
  \centering
  \small
  \setlength{\tabcolsep}{3.5pt}
  \begin{tabular}{@{}lrrr@{}}
    \toprule
    \textbf{Cadence feature} & \textbf{AI} & \textbf{Human} & \textbf{AUC} \\
    \midrule
    Articulation rate (w\,s$^{-1}$) & 3.68 & 2.51 & 0.844 \\
    Inter-word gap, CV              & 3.50 & 1.76 & 0.834 \\
    Inter-word gap, p90 (s)         & 0.000 & 0.138 & 0.687 \\
    Word duration, CV               & 0.373 & 0.447 & 0.706 \\
    \midrule
    \multicolumn{4}{@{}l}{\textit{fused, held out} ($n_{+}{=}67$, $n_{-}{=}62$, 200 splits)}\\
    ROC-AUC & \multicolumn{3}{r}{$\mathbf{0.952}$ \footnotesize[0.902, 0.990]} \\
    Recall at $5\%$ budget & \multicolumn{3}{r}{$0.774$} \\
    \midrule
    Voicemail control, AUC vs.\ human & \multicolumn{3}{r}{$0.707$} \\
    AI vs.\ voicemail, held out & \multicolumn{3}{r}{$0.895$ \footnotesize[0.848, 0.926]} \\
    \bottomrule
  \end{tabular}
  \vspace{-2mm}
  \caption{Far-end timing on matched production calls. The cadence separates recorded or synthesised speech from live conversation.}
  \label{tab:timing}
  \vspace{-4mm}
\end{table}

\subsection{All Four Classes Together}
\label{sec:inhouse}\label{sec:cascade}

Table~\ref{tab:inhouse} reports FEND end to end on the held-out half of the production calls, where
all four classes are labelled and the distribution is the unbalanced one the deployment sees. The
decision that gates the agent is solved. Separating a live human from any automated far end reaches F1 $0.979$, at precision $0.987$ and
recall $0.971$: of the $241$ live humans, $234$ keep the \textsc{human} label and $7$ are given an
automated one. Overall four-way accuracy is $0.938$. What
remains hard is telling the automations apart. AI callee recall is $0.457$ at precision $0.762$, so
when that detector survives arbitration it is usually right and its errors are missed detections,
not false alarms, with $18$ of $35$ AI calls assigned to voicemail. The confusion runs one way, and it runs in the expensive direction: treating an AI callee as
voicemail throws away a conversation that could still reach the callee, while the reverse only
wastes a message.

The window governs the human decision; one ordering relation governs the rest. On the public
evaluation half, accuracy rises from $94.3\%$ at a $5$ second window to $98.2\%$ at $15$ seconds
and is flat after, so a deployment that commits within the three to five seconds common in
practice is deciding before the evidence has arrived. Sweeping the priority order on the in-house half, the human-versus-automated decision is
invariant: F1 $0.979$, with exactly $7$ of $241$ live humans mislabelled under every ordering.
Among the automated classes one relation carries the result. Ranking voicemail above AI callee is
worth $14$ points of four-way accuracy, $0.938$ against $0.796$, because inverting it sends $88$
voicemail calls to the AI callee label rather than $18$ the other way, and AI callee precision
falls from $0.762$ to $0.260$. That is the relation conclusiveness prescribes: a record-now beep
or a menu prompt is near-decisive, while timing evidence fits three of the four classes. The
position of \textsc{ivr} moves accuracy by $0.2$ points (Appendix~\ref{app:cascade}).

\begin{table}[t]
  \centering
  \small
  \setlength{\tabcolsep}{4pt}
  \begin{tabular}{@{}lccccl@{}}
    \toprule
    \textbf{Class} & \textbf{n} & \textbf{Prec} & \textbf{Rec} & \textbf{F1} & \textbf{FP into} \\
    \midrule
    \textsc{human}  & 241 & 0.987 & 0.971 & \textbf{0.979} & IVR 1, AI 1 \\
    voicemail/AM    & 223 & 0.912 & 0.982 & \textbf{0.946} & AI 18, H 3 \\
    \textsc{ivr}    & 2   & 0.333 & 0.500 & 0.400 & \textit{n too small} \\
    AI callee       & 35  & 0.762 & 0.457 & 0.571 & VM 2, H 3 \\
    \midrule
    \textsc{human} vs rest & 241 & 0.987 & 0.971 & \textbf{0.979} & 7 humans \\
    \bottomrule
  \end{tabular}
  \vspace{-2mm}
  \caption{Per-class results on the held-out half ($n{=}501$) of the in-house calls, at the
  deployment's own class distribution, under a $20$\,s detection window. \textbf{FP into} names the
  true classes most often given this label. Overall accuracy $0.938$. Only the AI callee component
  is fitted on in-house data, on the other half. The \textsc{ivr} row rests on two calls and is
  reported for completeness only.}
  \label{tab:inhouse}
  \vspace{-4mm}
\end{table}

\subsection{Latency, Cost and Deployment Impact}
\label{sec:cost}

\paragraph{Latency and cost.} Table~\ref{tab:cost} reports time from call answer to firing on the
production calls. These are end-to-end decision times and not compute times: every detector that
reads text waits for the ASR, and the tone detector waits for the greeting to reach its beep.
Compute is negligible for three of the four, and only the LLM has a real per-call cost, which is
prompt tokens on a call the model is already making, since it is a tool on the model already
running the dialogue. The ordering is informative. The AI callee detector is the fastest at a
median of $2.02$ seconds because it needs only a few timed words, while the evidence that a mailbox
is speaking takes three to four times longer. The class our arbitration ranks last is therefore the
one whose evidence arrives first, which is why priority is by how conclusive the evidence is and
not by when it appears.

\vspace{-1mm}

\paragraph{Deployment considerations.} On the in-house mix, $51.9\%$ of answered calls are not a live person, so more than half of all
outbound attempts are decided by this component and not by the dialogue that follows. Of those,
$6.9$ points are AI callees, a class the binary label cannot express: under a human-or-machine
policy each is treated as a mailbox and the call ends, discarding the one automated far end that
can still route the agent to the person it called. In the other direction, HFPR bounds the damage
the system can do. At $2.9\%$, $7$ of the $241$ live people in the held-out half are given an automated label. This
is what makes a four-class detector usable on live traffic.

\begin{table}[t]
  \centering
  \small
  \setlength{\tabcolsep}{3.5pt}
  \begin{tabular}{@{}lccl@{}}
    \toprule
    \textbf{Detector} & \textbf{Med.} & \textbf{p90} & \textbf{Cost} \\
    \midrule
    Cadence (AI callee) & $2.02$\,s & $6.98$\,s  & negligible \\
    Keyphrase       & $3.24$\,s  & $14.70$\,s & negligible \\
    Tone (beep)     & $8.86$\,s  & $16.25$\,s & negligible \\
    LLM (voicemail) & $8.91$\,s  & $13.07$\,s & $604{+}41$ tok \\
    \bottomrule
  \end{tabular}
  \vspace{-2mm}
  \caption{Time from call answer to firing on the $1{,}000$ in-house calls, conditioned on firing
  within the $20$\,s window, with per-call cost. Token counts are median prompt and completion.}
  \label{tab:cost}
  \vspace{-4mm}
\end{table}

\vspace{-1mm}

\section{Conclusion}
\label{sec:conclusion}
\vspace{-1mm}

We treated the far end of an outbound Voice AI call as a four-class detection problem, evaluated
on a public benchmark and on $1{,}000$ hand-annotated production calls.
\textsc{voicemail/am} is solved from content, at $100\%$ accuracy on a held-out public split with
no false positives on humans, where the open-source baseline mislabels $30$ to $66\%$ of live
humans. \textsc{ivr} becomes separable once the class is defined by whom the automation speaks for.
The \textsc{ai callee} is the open case, and the class AMD has no label for. Its spectrum does not
identify it, since measurements that appear to work either fail a human-versus-human calibration
or track the recording channel; its speech timing does, at $0.952$ ROC-AUC. End to end, the
decision that gates the agent reaches F1 $0.979$, while telling the automations apart remains the
harder half. The fastest-growing far end is the one that must be detected from how it speaks, not from what it
says.

\newpage

\section*{Limitations}
\label{sec:limitations-c5}

\textbf{The in-house class mix is one deployment's.} The $1{,}000$ calls come from one outbound
programme, so the proportions describe its callee list, calling hours and country mix and not
outbound Voice AI in general. The \textsc{ivr} row of Table~\ref{tab:inhouse} therefore rests on two
calls and is reported only for completeness, and the AI callee rate of $6.9\%$ should be expected to
move quickly across markets, since consumer call screening is switched on by handset vendors and not
adopted call by call.

\textbf{Every AI callee we observe is a screener.} All AI callee calls in our sample are the same
subtype, and a gatekeeping assistant with a scripted opening. We have no measured instance of an AI receptionist, personal assistant or third-party agent
conducting a call, so the claim that the class will grow beyond screening rests on where the
technology is going and not on our data, and the timing model establishes only that one product can
be told apart.

\textbf{The acoustic result is negative, and tested only against read-prompt speech.} Our only
usable synthetic arm is VoiceBench, which is read from prompts. The arm that would have tested conversational synthesis over a live carrier failed a channel check
and was excluded, and the window normalisation retains $92\%$ of one human corpus but only $48\%$
of the synthetic one, leaving a selection difference between the classes compared. After
calibration one metric survives, so the $0.624$ AUC rests on a single statistic. No corpus of
genuinely far-end agent-answered calls exists, and building one is the necessary next step.

\textbf{The timing result depends on the deployment's ASR and on a talkative minority.} The $0.952$ AUC is measured on $67$ AI and $62$ human calls, the $13\%$ of people whose opening
turn carries at least four timed words. The timings come from the deployment's commercial ASR, so
a recogniser with different word-boundary behaviour could align synthesised speech differently,
and we have not reproduced the result on an open ASR.

\textbf{The comparisons are uneven.} The public corpus is short clips and probably
under-represents long greetings and human pickup after a greeting, which is why the four-class
evaluation runs on in-house traffic that is not public. Commercial AMD runs inside the carrier media
path and cannot be replayed, so those numbers stay published operating points. The closest learned
prior work~\citep{altwlkany2024rnn} releases no checkpoint, so we compare against a rule-based
system only, and the LLM detectors use a closed model whose pre-training data we cannot audit
(Appendix~\ref{app:contamination}).

\section*{Ethical Considerations}

The in-house calls contain customer content and are not released, all examples are anonymised, and
no personal data is reproduced. Outbound messaging should still respect applicable consent,
disclosure and do-not-call rules.

\paragraph{Deciding whether the other party is human.} The AI callee class asks a system to judge
whether a person is on the line, and the error is asymmetric morally as well as operationally, since
treating a human as a machine can mean talking over them, hanging up on them, or sending them down a
degraded path. That is why we report the operating point under a budget on human false positives and
treat a detector with zero recall at that budget as undeployable. The inference is also
adversarially exposed, because an operator who wants to avoid detection can change pacing at little
cost. We describe the features anyway, since a closed detector whose failure modes nobody has
measured is worse for the people on the receiving end.

\paragraph{Detection is not a substitute for disclosure.} EU AI Act Article~50 \citep{euaiact2024}
and California's B.O.T. Act \citep{botact2018} require automated callers to identify themselves.
Inferring what regulation says should be volunteered risks making non-disclosure feel normal. We
regard detection as a defensive complement to disclosure, not a reason for a calling agent to
disclose less.

\paragraph{Treating callers differently by class.} Where the inferred class is wrong, the difference
in treatment lands on a real person, and the far ends most likely to be misread, such as accented
speech or poor channels, are not evenly distributed across callees. Our public evaluation covers four languages and our human reference corpora are two call-centre
populations, so we cannot say how error rates vary across speaker groups, and that measurement is
a prerequisite for deploying class-based policy at scale. Relatedly, what an agent should do on
recognising another agent is a question about whose interests the call serves, which we leave as a
configured policy.

\newpage

\bibliography{custom}

\appendix

\section{Asterisk \texttt{app\_amd} Port}
\label{app:asterisk}
The baseline of Section~\ref{sec:baselines} is a line-for-line port of two Asterisk sources:
the \texttt{isAnsweringMachine} state machine of \texttt{apps/app\_amd.c}, which is a
threshold cascade over word and silence durations, and the silence detector of
\texttt{main/dsp.c}, which declares a frame silent when its mean absolute sample value falls
below \emph{silence threshold} and accumulates consecutive silent frames. We preserve the
order in which \texttt{app\_amd} tests its thresholds, its initial \textsc{in\_word} state
(so the first word of a greeting is not counted), and the fact that \emph{voice duration}
accumulates across the whole call while \emph{consecutive voice duration} resets on any
silence run reaching \emph{between words silence}.

Audio is resampled to $8$\,kHz mono signed-linear and framed at $20$\,ms ($160$ samples),
matching an RTP-fed channel. A recording ends but a connected call does not, so clips shorter
than \emph{total analysis time} are padded with digital silence, which is what a channel that
stays up delivers; $79$ of $400$ clips are affected at the shipped defaults. \texttt{app\_amd}
returns \textsc{machine}, \textsc{human} or \textsc{notsure}, and a deployment must map the
last of these to an action; we score both mappings and report the conservative one
(\textsc{notsure} $\rightarrow$ human, which protects live callers). It changes nothing
material here, because only one test clip returns \textsc{notsure} at the defaults and none
does under the tuned configuration.

The grid search covers \emph{initial silence} $\in\{1.5,2.5,4\}$\,s, \emph{greeting}
$\in\{1.5,2.5,4,6\}$\,s, \emph{after greeting silence} $\in\{0.4,0.8,1.5,2.5\}$\,s,
\emph{total analysis time} $\in\{5,10\}$\,s, \emph{maximum number of words}
$\in\{2,3,5,8\}$ and \emph{silence threshold} $\in\{128,256,512\}$: $1{,}152$ configurations.
The selected configuration raises every threshold well above Asterisk's defaults
($4$\,s initial silence, $4$\,s greeting, $2.5$\,s after-greeting silence, $10$\,s analysis,
$5$ words), which is the search telling us the defaults are tuned for a far end that answers
with a short ``Hello?'' rather than for the greeting lengths in this corpus.

\section{Dataset Composition}
\label{app:datasets}
Table~\ref{tab:datasets} gives the composition of the Telephony-AMD test split used
throughout Section~\ref{sec:experiments}.

\begin{table}[h]
  \centering
  \small
  \setlength{\tabcolsep}{4pt}
  \begin{tabular}{@{}lcccc@{}}
    \toprule
    & \textsc{human} & \textsc{vm} & \textsc{am} & \textsc{ivr} \\
    \midrule
    Clips & 114 & 102 & 92 & 92 \\
    \bottomrule
  \end{tabular}
  \caption{Telephony-AMD test split ($400$ clips): class composition (top) and
  ASR-detected language composition (bottom).}
  \label{tab:datasets}
\end{table}

The internal sets are: a beep \emph{golden set} of $293$ beep-positive clips
(tone $\sim$$350$--$1000$\,Hz) and $74$ beep-negative conversations; a \emph{beep-free} set of
$\sim$$65{,}000$ clips including hold music and filler speech, used only for false-alarm
validation; and a random sample of $1{,}000$ outbound in-house Voice AI calls drawn over a fixed
period, of which $478$ showed no AMD activity and $464$ were detected by at least one
detector. These cannot be released, for compliance and data-privacy reasons, as with the
proprietary corpora of prior work~\citep{altwlkany2024rnn,anisimov2017cnn}; the Telephony-AMD
evaluation is public and repeatable.

\section{Data Contamination}
\label{app:contamination}
The LLM detector is a general-purpose model prompted zero-shot and is not fine-tuned on any
AMD corpus, so contamination could only take the form of Telephony-AMD audio or transcripts
having entered pre-training. We cannot audit a closed model's corpus, but three observations
bound the concern. First, the detector is not asked to recall a benchmark label: it sees a streaming transcript and decides whether to emit a tool call, and is scored on
\emph{whether and when} it fires inside a live call, not on a held-out label. Second, its
behaviour on Telephony-AMD matches its behaviour on our in-house Voice AI traffic, which post-dates
the model and is not public, in both it is the sole signal on the same category of far end
(AI screeners, conversational greetings with no beep). Third, the two detectors carrying most
of the recall, beep and keyphrase, involve no learned model at all, so the headline OR-fusion
result does not rest on the LLM path.

\section{Behaviour on IVR}
\label{app:ivr}
IVR is a third far-end behaviour, not a mislabelled voicemail: the correct agent response is
to stay on the line and navigate the menu, never to leave a message and terminate. On the $92$
IVR clips of the Telephony-AMD test split the beep detector fires on $0$ ($0.0\%$), since menus
carry no record-now beep, and the keyphrase detector on $0$ ($0.0\%$), since no voicemail
phrase matches a menu prompt. The LLM detector fires on $7$ ($7.6\%$). Three of those seven are
genuinely ambiguous far ends, two hold queues that themselves offer a voicemail fallback
(\emph{``Votre appel est en attente\dots\ Pour laisser un message\dots''}) and one automated
business-hours announcement, and the remaining four fired on short ASR fragments
(\emph{``Your call\dots''}) and are prompt failures we count against the detector. A spurious IVR firing does not by itself end the call, but it does put the agent on the wrong
track. Tightening the prompt against fragmentary transcripts is future work.

\section{Beep Detector Details}
\label{app:beep}

\paragraph{What a beep looks like.} From manual inspection of collected beeps, the
tone is close to a single sinusoid: frequency $350$--$2000$\,Hz, duration
$0.5$--$2.0$\,s, one dominant spectral peak with few harmonics, and energy in a
narrow band. The hard cases are a beep masked by speech or environmental noise,
carrier frequency jitter, variable duration, codec artifacts, and tonal hold-music
that mimics a beep.

\paragraph{Four conditions.} A beep is declared only when \emph{all four} hold on
the same frequency bin, each closing one way a non-beep could pass:
\begin{enumerate}\setlength{\itemsep}{1pt}\setlength{\topsep}{2pt}\setlength{\parskip}{0pt}
  \item \textbf{Dominant peak:} the top peak exceeds the second by
  $\Delta\mathrm{dB}=20\log_{10}(M_1)-20\log_{10}(M_2)\ge 20$\,dB. A beep is one
  pure tone, whereas speech is broadband.
  \item \textbf{Energy isolation:} the mean energy just below ($E_{\text{low}}$)
  and just above ($E_{\text{high}}$) the peak are both under threshold. It rejects
  broadband noise and music, whose energy spreads across neighbouring bins.
  \item \textbf{Frequency floor:} the peak sits above $350$\,Hz . It rejects
  low-frequency hum, voiced filler, and DTMF tones.
  \item \textbf{Streak:} the peak persists for \emph{streak required} $=4$
  consecutive frames. It rejects transient tonal blips.
\end{enumerate}
The streaming pipeline is: PCM $\rightarrow$ float32 normalization $\rightarrow$
windowed STFT (Hann window~\citep{harris1978windows}, computed with the
FFT~\citep{cooley1965fft}) $\rightarrow$ peak detection (top bins, neighbour merge,
dominance check) $\rightarrow$ energy/frequency isolation $\rightarrow$ streak
validation $\rightarrow$ beep.

\paragraph{Parameters.} The STFT front-end is fixed by the telephony stream. The
\emph{detection thresholds} are what we tune (Table~\ref{tab:beepparams}). Only four
thresholds drive the decision which includes the dominant-peak margin, the energy-isolation
test, the streak length, and the frequency floor, and we set them on the golden
set below.

\begin{table}[h]
  \centering
  \small
  \setlength{\tabcolsep}{3pt}
  \begin{tabular}{@{}llp{3.15cm}@{}}
    \toprule
    \textbf{Parameter} & \textbf{Default} & \textbf{Role} \\
    \midrule
    \multicolumn{3}{@{}l}{\textit{STFT front-end}}\\
    sampling rate     & 8\,kHz      & telephony stream rate \\
    window / hop      & 64 / 32\,ms & freq.\ resolution \\
    FFT size          & 512         & spectral detail \\
    \midrule
    \multicolumn{3}{@{}l}{\textit{Detection thresholds (tuned on golden set)}}\\
    dominant-peak margin & 20\,dB   & $M_1$ vs.\ $M_2$ selectivity \\
    energy isolation     & on       & reject broadband or music \\
    streak required      & 4        & consecutive frames \\
    frequency floor      & 350\,Hz  & reject low hum \\
    min magnitude        & 0.1      & ignore low-energy noise \\
    \bottomrule
  \end{tabular}
  \caption{Beep detector parameters. The STFT front-end is fixed by the $8$\,kHz
  telephony stream. The four detection thresholds are tuned on the golden set.}
  \label{tab:beepparams}
\end{table}

\begin{figure}[h]
  \centering
  \includegraphics[width=\columnwidth]{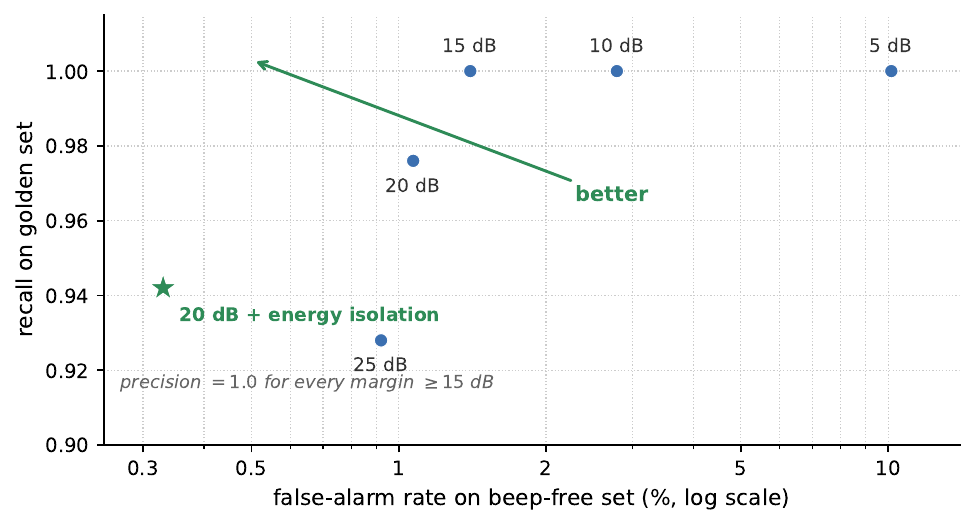}
  \caption{Beep operating-point selection: recall (golden set) vs.\ false-alarm
  rate on the beep-free set as the dominant-peak margin varies (streak $=4$;
  log $x$-axis). The chosen $20$\,dB $+$ energy-isolation config (star) gives the
  lowest false-alarm rate at near-top recall, looser margins raise false alarms,
  tighter ones cost recall. Precision is $1.0$ for every margin $\ge15$\,dB.}
  \label{fig:beeptradeoff}
\end{figure}

\paragraph{Choosing the operating point.} We tune on the internal golden set
($293$ beep-positive, $74$ negative) and \emph{validate false alarms} on a large
beep-free set ($\sim$$65$k clips), since a spurious beep would wrongly trigger AMD
on a live call. Two targets drive the choice: precision must be $1.0$ (no false
alarms on golden), and the false-alarm rate on the beep-free stream must be as low
as possible. Figure~\ref{fig:beeptradeoff} sweeps the dominant-peak margin. A loose
margin ($5$--$10$\,dB) catches every beep (recall $1.0$) but false-alarms on up to
$10\%$ of beep-free clips, $15$\,dB still leaves $1.4\%$. Adding the
energy-isolation test at $20$\,dB drops the false-alarm rate to $0.33\%$ at
precision $1.0$ and recall $0.942$ (missing $17/293$ positives, all very-low-energy
or noise-masked beeps),
while $25$\,dB costs more recall for no gain. So \textbf{$20$\,dB $+$ energy
isolation is the knee}, and our default.

\paragraph{Why streak $=4$.} The streak length trades transient rejection against
recall and latency. A streak of $3$ keeps recall high but stays trigger-happy
(false-alarm rate $\sim$$1.0$--$1.4\%$); a streak of $5$ over-suppresses (golden
recall falls toward $0.75$ at tight frequency tolerance) and adds latency. A streak
of $4$ balances both, cutting the false-alarm rate to $0.44\%$ for a modest
recall cost at only $4\times32\,\text{ms}\approx128$\,ms of confirmation delay. The
internal sets cannot be released, for compliance and data-privacy reasons.

\section{Voicemail Keyphrases}
\label{app:keyphrases}
The keyphrase detector matches the normalized ASR transcript against a
configurable phrase list. We derived the English list from recurring greetings in
internal answering-machine call transcripts. For the Telephony-AMD experiment,
which spans English, Spanish, French, and German, we added translations of these
phrases into the other three languages.

\begin{tcolorbox}[
  colback=kpc!6, colframe=kpc!75!black,
  boxrule=0.6pt, arc=2pt, left=6pt, right=6pt, top=5pt, bottom=5pt,
  title=\textbf{English keyphrases}, fonttitle=\small\bfseries,
  coltitle=white, colbacktitle=kpc!75!black]
\small\itshape
\setlength{\parskip}{1.5pt}\setlength{\parindent}{0pt}
please leave your name \\
forwarded to voicemail \\
record your message \\
at the tone \\
after the tone \\
after the beep \\
can't come to the phone \\
voice messaging system \\
automated voice message \\
has been forwarded to \\
when you've finished recording \\
when you have finished recording \\
hang up or press \\
unavailable right now
\end{tcolorbox}

\begin{tcolorbox}[
  colback=kpc!6, colframe=kpc!75!black,
  boxrule=0.6pt, arc=2pt, left=6pt, right=6pt, top=5pt, bottom=5pt,
  title=\textbf{Spanish keyphrases}, fonttitle=\small\bfseries,
  coltitle=white, colbacktitle=kpc!75!black]
\small\itshape
\setlength{\parskip}{1.5pt}\setlength{\parindent}{0pt}
por favor deje su nombre \\
desviado al buzón de voz \\
grabe su mensaje \\
después de la señal \\
después del tono \\
después del bip \\
no puede atender el teléfono \\
sistema de mensajería de voz \\
mensaje de voz automático \\
ha sido desviado a \\
cuando haya terminado de grabar \\
cuelgue o presione \\
no está disponible en este momento
\end{tcolorbox}

\begin{tcolorbox}[
  colback=kpc!6, colframe=kpc!75!black,
  boxrule=0.6pt, arc=2pt, left=6pt, right=6pt, top=5pt, bottom=5pt,
  title=\textbf{French keyphrases}, fonttitle=\small\bfseries,
  coltitle=white, colbacktitle=kpc!75!black]
\small\itshape
\setlength{\parskip}{1.5pt}\setlength{\parindent}{0pt}
veuillez laisser votre nom \\
transféré vers la messagerie vocale \\
enregistrez votre message \\
au signal sonore \\
après la tonalité \\
après le bip \\
ne peut pas répondre au téléphone \\
système de messagerie vocale \\
message vocal automatique \\
a été transféré vers \\
lorsque vous avez terminé l'enregistrement \\
raccrochez ou appuyez \\
n'est pas disponible pour le moment
\end{tcolorbox}

\begin{tcolorbox}[
  colback=kpc!6, colframe=kpc!75!black,
  boxrule=0.6pt, arc=2pt, left=6pt, right=6pt, top=5pt, bottom=5pt,
  title=\textbf{German keyphrases}, fonttitle=\small\bfseries,
  coltitle=white, colbacktitle=kpc!75!black]
\small\itshape
\setlength{\parskip}{1.5pt}\setlength{\parindent}{0pt}
bitte hinterlassen Sie Ihren Namen \\
zur Mailbox weitergeleitet \\
nehmen Sie Ihre Nachricht auf \\
nach dem Signalton \\
nach dem Ton \\
nach dem Piepton \\
ist gerade nicht erreichbar \\
Sprachnachrichtensystem \\
automatische Sprachnachricht \\
wurde weitergeleitet an \\
wenn Sie mit der Aufnahme fertig sind \\
legen Sie auf oder drücken Sie \\
im Moment nicht erreichbar
\end{tcolorbox}

\newpage
\section{Voicemail Detector Prompt}
\label{app:llmprompt}\label{app:amdprompt}

\begin{tcolorbox}[
  colback=llmc!6, colframe=llmc!75!black,
  boxrule=0.6pt, arc=2pt, left=6pt, right=6pt, top=5pt, bottom=5pt,
  title=\textbf{Answering-machine detection prompt (frozen)}, fonttitle=\small\bfseries,
  coltitle=white, colbacktitle=llmc!75!black]
\small
\setlength{\parskip}{3pt}\setlength{\parindent}{0pt}

The LLM detector runs over the live transcript and raises the \emph{AMD pending} signal when the
far end shows automated-mailbox characteristics. The prompt below is the frozen version used for
every voicemail number reported in the paper. It differs from the one shipped in our production
pipeline in one respect, argued in Section~\ref{sec:results-amd}: the positive class is
\emph{any automated mailbox system answering on the callee's behalf}, not only one that invites a
recording. The detector additionally returns which state it observed, so that the downstream action can
separate ``retry later'' from ``stop calling this number'' without collapsing the two at detection
time. Both prompts are reproduced verbatim except that personal and company names inside the
illustrative examples are shown as $\langle$name$\rangle$ and $\langle$company$\rangle$.

You classify the callee side of an outbound phone call for answering-machine detection (AMD).

Call the answering\_machine\_detector tool whenever an automated voicemail, answering machine, or
carrier mailbox system has answered in place of the person being called. The decisive question is
whether a recorded or synthetic system answered on the callee's behalf, not whether a message
can actually be left right now.

Set \texttt{state} to whichever applies:
\begin{enumerate}\setlength{\itemsep}{1pt}\setlength{\parskip}{0pt}\setlength{\topsep}{2pt}
\item \texttt{greeting\_invites\_message}, the recording invites a message: a personal greeting (\emph{``you've reached $\langle$name$\rangle$, leave a message after the
tone''}), a carrier voicemail greeting naming a network or subscriber, or any instruction to record after a tone, beep or
  signal.
\item \texttt{mailbox\_unavailable}, an automated message reporting that the mailbox or line
  cannot take the call, so no recording is possible: the mailbox is full, has not been set up, or
  the subscriber is temporarily unreachable and the caller is asked to try later.
\item \texttt{number\_invalid}, a carrier intercept reporting that the number itself is
  unusable: not in service, no longer assigned, or the call cannot be completed as dialled.
\end{enumerate}
These messages occur in any language. Classify by what the message does, not by matching English
wording.

Do NOT call the tool when:
\begin{itemize}\setlength{\itemsep}{1pt}\setlength{\parskip}{0pt}\setlength{\topsep}{2pt}
\item A live person answers conversationally, greets the caller, asks a question, or engages in
  two-way dialogue, including brief or abrupt answers
\item The far end is an IVR menu, a queue or hold announcement, or a recorded business greeting
  that routes the call rather than taking a message. If a recording offers menu options and also
  mentions leaving a message, treat the menu as dominant and do NOT call the tool
\item The transcript is empty, untranscribed, or too garbled to identify
\end{itemize}
When the transcript is genuinely ambiguous between a live person and a recording, do NOT call the
tool.
\end{tcolorbox}

\section{IVR Detector Prompt}
\label{app:ivrprompt}
The IVR detector is a second tool exposed to the conversational LLM, distinguished from the
voicemail detector by what it is asked to look for. Where the voicemail prompt keys on a mailbox
system answering for the callee, this one keys on an IVR navigation: a menu, queue
or recorded greeting is a far end the agent should stay on and traverse, not one it should record
against.

\begin{tcolorbox}[
  colback=llmc!6, colframe=llmc!75!black,
  boxrule=0.6pt, arc=2pt, left=6pt, right=6pt, top=5pt, bottom=5pt,
  title=\textbf{IVR detection prompt}, fonttitle=\small\bfseries,
  coltitle=white, colbacktitle=llmc!75!black]
\small
\setlength{\parskip}{3pt}\setlength{\parindent}{0pt}

You classify the callee side of an outbound phone call for interactive voice response (IVR)
detection.

Call the IVR detector tool when the transcript shows any automated, pre-recorded or synthesised
system message from the callee's organisation:
\begin{itemize}\setlength{\itemsep}{1pt}\setlength{\parskip}{0pt}\setlength{\topsep}{2pt}
\item A menu offering choices (\emph{``press one for\dots''}, \emph{``say sales or say
  support''})
\item An instruction to enter digits (\emph{``enter your account number''}, \emph{``followed by
  the pound key''})
\item A queue or hold announcement (\emph{``your call will be answered in the order it was
  received''}, \emph{``please stay on the line''}, \emph{``estimated wait time''})
\item A recorded greeting or business-hours announcement (\emph{``thank you for calling
$\langle$company$\rangle$''}, \emph{``our offices are currently closed''})
\item A monitoring or recording notice delivered as a recorded system message, or a
  language-selection prompt
\end{itemize}
The decisive question is whether a recorded or synthetic system is speaking on behalf of an
\textsc{organisation}, not whether a menu is offered: a greeting or hold message with no
selectable option still counts, provided it is an organisation's system answering its own line.

Do NOT call the tool when:
\begin{itemize}\setlength{\itemsep}{1pt}\setlength{\parskip}{0pt}\setlength{\topsep}{2pt}
\item The far end is an assistant answering for an \emph{individual} rather than an
  organisation: it offers to take the caller's name and reason and see whether that person is
  available, or asks the caller to hold while it fetches them. That is a personal call screener
  (class~4), not an IVR, even though it is automated and may ask the caller to stay on the line
\item The far end invites the caller to leave a recorded message, that is voicemail
\item A live person answers conversationally, even if they say \emph{``thank you for calling''},
  give a company name, or mention recording
\item The transcript is empty, untranscribed, or too garbled to tell
\item You are uncertain, when in doubt, do NOT call the tool
\end{itemize}
\end{tcolorbox}

\paragraph{Tuning.} The prompt was iterated on the $119$-clip tuning split of
Section~\ref{sec:datasets} only; the $281$-clip evaluation split was not inspected during
prompt development.

\section{Signal-Layer Panel}
\label{app:signal}

The panel has five metrics, each a summary of the far-end signal over the normalised window.
\emph{Formant-band concentration} is the fraction of spectral energy in $200$--$800$\,Hz, where the
first formants of voiced speech sit. \emph{Voicing continuity} is the fraction of frames judged
voiced by a normalised-autocorrelation pitch test. \emph{Level dynamics} is the mean frame RMS over
$50$\,ms windows. \emph{Background flatness} is spectral flatness, the ratio of the geometric to the
arithmetic mean of the spectrum, which is high for noise-like audio and low for tonal audio.
\emph{Segment structure} is the mean duration of a contiguous speech segment under an energy-gated
VAD.

Each is calibrated on human speech alone, by the rule of Section~\ref{sec:setup}, before being
compared against synthesis. Table~\ref{tab:asfs} reports two null values per metric. The
\emph{within} column is the $99$th percentile of $|g|$ over $200$ random splits of a single human
corpus, which is the noise floor: how large $|g|$ gets when there is no real difference. The
\emph{cross} column is $|g|$ between the two independent human corpora, which should stay at that
floor if the metric measures the speaker rather than the corpus. A metric is admitted only when the
cross value does, at $|g|\le 1$. Four fail, each by a wide margin, and all four correlate with
realised speech fraction while the survivor does not.

The surviving metric is then fitted on a training half and read once on a held-out half split by
source call, with the threshold set as described in Appendix~\ref{app:oppoint}. It reaches $0.624$ ROC-AUC and has no usable recall at a $1$--$2\%$ false-positive
budget. The robocall control, which is pre-recorded \emph{human} speech, scores below the human
negatives ($0.414$), so the metric is weakly specific to synthesis and not merely to audio that is
not a live call. It is still too weak to act on.

\begin{table}[h]
  \centering
  \small
  \setlength{\tabcolsep}{3pt}
  \begin{tabular}{@{}lrrr@{}}
    \toprule
    \multicolumn{4}{@{}l}{\textit{null calibration, train only} ($|g|$)}\\
    \textbf{Metric} & \textbf{within p99} & \textbf{cross} & \textbf{adm.} \\
    \midrule
    Formant-band conc.  & 0.25 & 1.24 & no \\
    Voicing continuity  & 0.25 & 2.79 & no \\
    Level dynamics      & 0.21 & 1.11 & no \\
    Background flatness & 0.28 & 3.95 & no \\
    Segment structure   & 0.23 & \textbf{0.26} & \textbf{yes} \\
    \midrule
    \multicolumn{4}{@{}l}{\textit{held-out test} ($n_{+}{=}257$, $n_{-}{=}1324$)}\\
    ROC-AUC vs.\ human  & \multicolumn{3}{r}{$0.624$ \footnotesize[0.597, 0.652]} \\
    \midrule
    \multicolumn{4}{@{}l}{\textit{operating points, threshold fitted on train}}\\
    \textbf{FPR budget} & \textbf{test FPR} & \textbf{recall} & \textbf{FP} \\
    $1\%$  & 1.06\% & \textbf{0.000} & 14 \\
    $2\%$  & 1.74\% & \textbf{0.000} & 23 \\
    $5\%$  & 3.25\% & 0.004 & 43 \\
    \midrule
    Robocall control, AUC vs.\ human & \multicolumn{3}{r}{$0.414$} \\
    \bottomrule
  \end{tabular}
  \caption{Signal-layer panel with a held-out threshold (Section~\ref{sec:aicallee}).}
  \label{tab:asfs}
\end{table}

\section{Arbitration Sweep}
\label{app:cascade}

The priority order is swept on the in-house held-out half, which carries all four classes; the
public split cannot exercise it, because it contains no AI callee and so cannot produce the
conflicts the ordering exists to resolve. Table~\ref{tab:cascade-inhouse} reports all six orderings
of the three automated classes at a $20$\,s window ($n{=}501$).

Two things hold across the sweep. The human-versus-automated decision does not move at all: F1
$0.979$, precision $0.987$, recall $0.971$, and exactly $7$ of $241$ live humans given an automated
label under every ordering. And the ordering matters only through the voicemail-versus-AI-callee
relation. With voicemail ranked first, four-way accuracy is $0.936$ to $0.938$; invert it and
accuracy falls to $0.796$ to $0.798$, as $88$ voicemail calls are read as AI callee rather than $18$
AI callees read as voicemail. Voicemail recall drops from $0.982$ to about $0.59$ and AI callee
precision from $0.762$ to $0.260$. The position of \textsc{ivr} is immaterial at $0.2$ points, which
is what two calls can support.

Table~\ref{tab:cascade} sweeps the decision window and the ordering on the public evaluation half,
where the window effect is measurable: it moves accuracy by $3.9$ points. The ordering is
uninformative there for the reason above, and in any case only three of $281$ clips trigger two
components, with one of those firing both at the same timestamp to the hundredth of a second, so
``first'' is decided by implementation order. We report those policies as equivalent on that split
and draw the ordering conclusion from the in-house sweep instead.

\begin{table}[h]
  \centering
  \small
  \setlength{\tabcolsep}{3.5pt}
  \begin{tabular}{@{}lccccc@{}}
    \toprule
    \textbf{Priority order} & \textbf{H F1} & \textbf{H FP} & \textbf{VM F1} & \textbf{AI F1} & \textbf{Acc} \\
    \midrule
    \textsc{vm}\,$>$\,\textsc{ivr}\,$>$\,\textsc{ai} & 0.979 & 7 & \textbf{0.946} & \textbf{0.571} & \textbf{0.938} \\
    \textsc{vm}\,$>$\,\textsc{ai}\,$>$\,\textsc{ivr} & 0.979 & 7 & \textbf{0.946} & \textbf{0.571} & \textbf{0.938} \\
    \textsc{ivr}\,$>$\,\textsc{vm}\,$>$\,\textsc{ai} & 0.979 & 7 & 0.944 & \textbf{0.571} & 0.936 \\
    \midrule
    \textsc{ai}\,$>$\,\textsc{vm}\,$>$\,\textsc{ivr} & 0.979 & 7 & 0.737 & 0.405 & 0.798 \\
    \textsc{ai}\,$>$\,\textsc{ivr}\,$>$\,\textsc{vm} & 0.979 & 7 & 0.733 & 0.405 & 0.796 \\
    \textsc{ivr}\,$>$\,\textsc{ai}\,$>$\,\textsc{vm} & 0.979 & 7 & 0.733 & 0.405 & 0.796 \\
    \bottomrule
  \end{tabular}
  \vspace{-2mm}
  \caption{Priority sweep on the in-house held-out half ($n{=}501$, $20$\,s window). \textbf{H FP}
  counts live humans given an automated label, out of $241$. \textsc{human} is the default
  throughout and is never outranked. The human decision is invariant; the automated classes turn on
  whether voicemail outranks AI callee.}
  \label{tab:cascade-inhouse}
  \vspace{-4mm}
\end{table}

\begin{table}[h]
  \centering
  \small
  \setlength{\tabcolsep}{3pt}
  \begin{tabular}{@{}llcccc@{}}
    \toprule
    \textbf{Order} & \textbf{$W$} & \textbf{Acc} & \textbf{Prec$_H$} & \textbf{Rec$_H$} & \textbf{FP$_H$} \\
    \midrule
    VM $\rightarrow$ IVR $\rightarrow$ AI & 5\,s  & 94.3 & 0.848 & 1.000 & 0 \\
    VM $\rightarrow$ IVR $\rightarrow$ AI & 15\,s & 98.2 & 0.963 & 1.000 & 0 \\
    VM $\rightarrow$ IVR $\rightarrow$ AI & 30\,s & 98.2 & 0.963 & 1.000 & 0 \\
    \midrule
    IVR $\rightarrow$ VM $\rightarrow$ AI & 30\,s & 98.6 & 0.963 & 1.000 & 0 \\
    \textit{first-firing (no priority)}   & 30\,s & 98.9 & 0.963 & 1.000 & 0 \\
    \bottomrule
  \end{tabular}
  \caption{Arbitration on the public evaluation half. \textbf{FP$_H$} counts live humans given an
  automated label. No policy mislabels a human at any window.}
  \label{tab:cascade}
\end{table}

\section{Cadence Classifier}
\label{app:cadence}

The AI callee detector is a logistic-regression classifier over seven timing features taken from
the far end's first turn: the median, standard deviation, coefficient of variation and $90$th
percentile of the gaps between consecutive words; the articulation rate in words per second; and
the median and coefficient of variation of word durations. Word start and end times come from the
deployment's ASR. Features are standardised on the training half, and the weights are fitted there
by full-batch gradient descent on the logistic loss ($4{,}000$ steps, learning rate $0.08$), with
\textsc{ai callee} as the positive class and \textsc{human} as the negative.

The decision threshold is set on the training half at a $5\%$ false-positive budget against the
human negatives, and is applied unchanged to the held-out half. A call whose opening far-end turn
carries fewer than four timed words receives no score and cannot fire: absence of evidence, which
is the correct behaviour for a component that reads timing, and the reason the negatives in
Table~\ref{tab:timing} are the $62$ of $481$ human far ends whose opening turn is long enough to
score. Table~\ref{tab:timing} reports individual ROC-AUCs for four of the seven features; the
fused model uses all seven.

\section{Signal-Layer Operating Point}
\label{app:oppoint}
The threshold for the signal-layer detector is fitted on the training half and applied unchanged
to the held-out half. Clips are split $40/60$ by source call, stratified by corpus and, for the
AppTek arm, by accent group, so that no call contributes to both halves and the accent mix is
matched. Training comprises $533$ Harper Valley, $348$ AppTek, $172$ VoiceBench and $225$ robocall
clips; test comprises $800$, $524$, $257$ and $338$ respectively.

For a target human false-positive budget $b$, the threshold is the $\lfloor b \cdot N_{-}\rfloor$-th
highest fused score among \emph{training} human clips. That threshold is then applied to the test
half without adjustment, and Table~\ref{tab:asfs} reports the realised test FPR beside the budget
it was set to; the two agree to within a point in every case, which is the check that the training
human score distribution transfers. Recall is the fraction of test TTS clips scoring above the same
threshold.

We report budgets of $1\%$, $2\%$ and $5\%$ and deliberately omit tighter ones. With $1{,}324$
human clips at test, a $0.5\%$ budget corresponds to roughly six admissible false positives, and
the sampling interval on a rate estimated from six events is wider than the differences between the
operating points being compared. A tighter budget can be quoted only with an order of magnitude
more human data, which is the practical reason the human arms, not the positive arm, are the ones
worth growing.

\end{document}